\documentclass[letterpaper, 10 pt, conference]{ieeeconf}  % Comment this line out if you need a4paper

\IEEEoverridecommandlockouts                              % This command is only needed if 
\usepackage{header} %libraries

\title{\LARGE \bf Transfer Learning for HVAC System Fault Detection}

\author{Chase P. Dowling and Baosen Zhang% <-this % stops a space
\thanks{Department of Electrical and Computer Engineering, University of Washington Seattle, WA 98195, USA	 \{cdowling, zhangbao\}@uw.edu}%
\thanks{This work is partially supported by grants from Centrica, plc. Real building data has been provided by the US Department of Energy's Pacific Northwest National Laboratory.}%
}

\begin{document}
\maketitle
\thispagestyle{empty}
\pagestyle{empty}

\begin{abstract}

Faults in HVAC systems degrade thermal comfort and energy efficiency in buildings and have received significant attention from the research community, with data driven methods gaining in popularity. Yet the lack of labeled data, such as normal versus faulty operational status, has slowed the application of machine learning to HVAC systems. In addition, for any particular building, there may be an insufficient number of observed faults over a reasonable amount of time for training. To overcome these challenges, we present a transfer methodology for a novel Bayesian classifier designed to distinguish between normal operations and faulty operations. The key is to train this classifier on a building with a large amount of sensor and fault data (for example, via simulation or standard test data) then transfer the classifier to a new building using a small amount of normal operations data from the new building.  We demonstrate a proof-of-concept for transferring a classifier between architecturally similar buildings in different climates and show few samples are required to maintain classification precision and recall.

\end{abstract}

%\keywords{transfer learning, HVAC, fault detection, kernel regression}

%\thanks{This work is partially supported by grants from Centrica, plc.}%

\maketitle

\section{INTRODUCTION}

\label{sec:intro}

Buildings account for roughly 40\% of electrical demand in the United States \cite{cao2016building} and climate control is one of the largest sources of power consumption in many buildings. The normal operation of heat, venting, and air conditioning (HVAC) systems is therefore critical for simultaneously maintaining energy efficiency and thermal comfort. Because of the widespread deployment of sensors, multiple data-driven algorithms have consequently been developed to detect faulty operation of HVAC systems~\cite{van2017advanced,li2016fault,perera2019machine}. A fundamental challenge of applying these types of machine learning algorithms, however, is the lack of \emph{labeled} data. 

Data-driven fault detection algorithms rely on having data about both the normal and faulted operation of HVAC systems~\cite{li2016fault, west2011automated}. Despite the growth in buildings equipped with a large number of sensors that can generate high resolution measurements~\cite{depuru2011smart}, it is nontrivial and labor intensive to correctly label each data point as coming from faulty or normal operation. In addition, HVAC systems mostly operate (fortunately) under fairly normal conditions. Therefore operators may need to either 1) wait for a long time to collect enough fault data---even if they can be correctly labeled---to train a useful algorithm, or 2) rely on established industry standards \cite{wen2012rp} and exhaustively simulate potential scenarios. Neither option takes direct and immediate advantage of the rich stream of data from well-equipped buildings.

Transfer learning \cite{long2017deep} provides a potential solution to these challenges. A predictor trained on an existing, labeled data set can be used as a starting point to train a predictor for the same task in which labeled data is limited but known to be in a similar embedding \cite{ling2008spectral}. Transfer learning has been used successfully in image classification \cite{raina2007self, zoph2018learning}. An image classifier, for example, is trained to recognize a set of image labels; to transfer the classifier to new images, initializing with the previously learned classifier requires far fewer examples of the new labels to achieve good accuracy. Transfer learning has only very recently begun to be applied to, for example, predicting energy consumption in buildings \cite{perera2019machine, singaravel2018deep}.

In this work we develop a \emph{transferable, naive Bayesian framework} for detecting faults and failures resulting from component degradation in three key steps:
\begin{enumerate}
    \item \emph{We derive a novel log-likelihood classifier that depends only on building normal operations data and an estimated state transition matrix}
    \item\emph{For a building with a large, labeled data set of HVAC component operations and weather data, we learn a normal operations state transition matrix}
    \item \emph{With the same model parameters, we transfer the learned state transition matrix with a limited number of samples from a similar building}
\end{enumerate}

We accomplish item (1) by specifying a matrix normal prior to derive a novel log-likelihood classifier that determines whether a series of HVAC system state observations was generated by the learned state transition matrix or by some other faulty state transition matrix having arisen as a random perturbation. Item (3) employs weighted least squares to transfer the learned state transition matrix to a new building for which labeled data is limited but the feature space is similar---e.g. type and number of relevant HVAC components---using model parameters learned in item (2).

To test our framework we first perform simple, motivating numerical simulations. We then proceed to transfer an hourly state transition model trained on a standard medium office building simulated by EnergyPlus~\cite{crawley2001energyplus} to a physically monitored \cite{haack2013volttron} testbed site, the Systems Engineering Building (SEB) located at Pacific Northwest National Laboratory (PNNL) \cite{follum2018online,goyal2019design} in Richland, WA. We separately compare how effectively a classifier can be transferred between similar office buildings simulated by EnergyPlus in different climates subject to known fault conditions. In Sec.~\ref{sec:model} we introduce the model and Sec.~\ref{sec:detection} describes the classifier and transfer procedure. In Sec.~\ref{sec:results} we present our results, and we conclude with Sec.~\ref{sec:conclusion}.

\section{MODEL}
\label{sec:model}
The state of an HVAC system can be described by a linear transformation with a polynomial kernel $\phi_{d}$~\cite{brown2012kernel}. Let $x_{t} \in \mathbb{R}^{n}$ be a dependent state variable (e.g. fan power, indoor temperature, pump status) and  $u_{t} \in \mathbb{R}^{k}$ be a independent state variable (e.g. time, outdoor temperature, humidity). Furthermore, let $\phi_{d}\left(x\right) = [x^{d}, x^{d-1},\ldots, x^{1}, \boldsymbol{1} ]$---the vector $x$ component-wise raised to the $i$'th power for $i \in [0,d]$ and concatenated---and $s_{t} = \left[ \phi_{d}\left(x_{t}\right), \phi_{d}\left(u_{t}\right) \right]^{T}$ the kernelized, concatenated dependent and independent state vectors of the HVAC system of dimension $p:= d(n + k)$. We will assume that if an HVAC system is operating normally, then a finite sequence of observations of $\left(s_{t}, x_{t+1}\right)$ will have been generated by
\begin{equation}\label{eqn:statetransition}
    x_{t+1} = As_{t} + \epsilon_{t}
\end{equation}
\noindent where $\epsilon_{t} \overset{\tiny iid}{\sim} \mathcal{N}(0,I)$ and $A \in \mathbb{R}^{n \times p}$ is the true state transition matrix, which can be estimated via kernel regression. A primary advantage of using kernel regression is that the state estimator $A$ is readily interpretable and easy to use in transfer learning, while in general also requiring fewer samples to parameterize and train than a more general model like a neural network \cite{brown2012kernel,singaravel2018deep,wei2017deep}. 

\section{Fault Detection and Model Transfer} \label{sec:detection}
\subsection{Bayesian Fault and Degredation Detection}
An entry $a_{i,j}$ of $A$ determines the relationship between the input and output states for explicit components of an HVAC system directly---we leverage this to derive a Bayesian classifier for determining if an HVAC system is operating in a faulty state without assuming explicitly what the faulty state transition matrix should look like.

By a faulty state, we mean that the HVAC is governed by some other transition matrix, denoted $\tilde{A}$. For an observation of $\left(s_{t}, x_{t+1}\right)$, there are two distinct probabilities: either the current dependent state of the HVAC system $x_{t+1}$ was generated by the normal state transition matrix $A$, or by some faulty state transition matrix $\tilde{A}$.

To compute these probabilities and derive a classifier, we make two assumptions: 1) we assume a Gaussian prior probability on the entries $\tilde{A}$ such that $\tilde{a_{ij}} \sim \mathcal{N}(a_{ij},1)$ (i.e. a matrix normal distribution \cite{gupta2018matrix} centered at $A$) and 2) initially, an HVAC system is equally likely to be operating in a faulty state or a normal state at any given time. Note these assumptions are used to derive the detection algorithm and provide some intuition, but the actual simulation study in Sec.~\ref{sec:results} uses real data which may not follow them. The probabilities of observing $x_{t+1}$ conditioned on the state transition matrices $A$ and $\tilde{A}$ are
\begin{subequations}
    \begin{align}
        P( x_{t+1} | A, s_{t} ) &= \frac{1}{(2\pi)^{n/2}}e^{-\frac{1}{2}\left[ (x_{t+1} - As_{t})^{T}(x_{t+1} - As_{t}) \right]} \label{eqn:nofaultprob}\\
        P( x_{t+1} | \tilde{A}, s_{t} ) &= \int \frac{1}{(2\pi)^{n/2}}e^{-\frac{1}{2}\left[ (x_{t+1} - Bs_{t})^{T}(x_{t+1} - Bs_{t}) \right]} \nonumber \\
        &\qquad \boldsymbol{\cdot} \frac{1}{(2\pi)^{np/2}}e^{-\frac{1}{2}\left[ (B - A)^{T}(B - A) \right]} \partial B. \label{eqn:posterior}
    \end{align}
\end{subequations}

Eqn.~\eqref{eqn:posterior} is the probability a sample was generated by some other state transition matrix $\widetilde{A} \neq A$ assumed to have a Bayesian prior probability distribution of a \emph{matrix normal random variable} with mean $A$ and identity row- and column-wise covariance (explained in Sec.~\ref{sec:results}). This approach fundamentally differs from many HVAC fault detection systems are rule-based \cite{schein2006hierarchical, wen2012rp}, or from algorithms that are learned from rule-based scenario generation \cite{wall2011dynamic}, as no modes of failure are assumed beforehand. These methods can be combined to produce more informative priors for entries of $A$ that describe the relationship between fan power consumption and air flow volume, for example. Furthermore, directly incorporating failure statistics with informed HVAC component failure rates \cite{hale2000survey} is also possible but outside the scope of this paper as we seek to \emph{naively} transfer a state transition matrix. By naive, we mean that no ground truth fault data is required to train. Once these probabilities are computed we can state a simple classification rule as 
\begin{equation}\label{eqn:classrule}
    0 \lessgtr \log [ P( x_{t+1} | A, s_{t} ) ] - \log [ P( x_{t+1} | \tilde{A}, s_{t} ) ],
\end{equation}
\noindent where a positive difference in log-likelihoods indicates the system is operating normally and a negative value indicates the observed data was generated by faulty operations. 

We show that \eqref{eqn:classrule} depends \textbf{only} on the entries of the normal operations state transition matrix $A$ and the sequence of observations $(s_{t}, x_{t+1})$. Indeed, for a single point $(s,x)$, the classification rule \eqref{eqn:classrule} simplifies to,
%\begin{subequations}
\begin{align}
    0 &\lessgtr Tr\left[ (x - As)^{T}(x - As)\right] - \nonumber \\
    &\qquad Tr\left[xx^{T} + AA^{T} - C^{-1}D^{T}D \right] - p\log(|C^{-1}|) \label{eqn:classeval} 
\end{align}
%\end{subequations}
where 
\begin{subequations}
\begin{align}
    C &:= (ss^{T} + I) \label{eqn:C}\\
    D &:= (xs^{T} + A) \label{eqn:D}
\end{align}
\end{subequations}
For convenience, let us define the binary classification output of \eqref{eqn:classeval} to be the function $\mathcal{C}: (s, x, A) \rightarrow \{0,1\}$.

%By our IID assumptions of the system noise $\epsilon_{t}$, the joint log-likelihood of multiple observations is additive, and thus multiple observations can be used to increase the general accuracy of the classification rule.

\subsection{Model Transfer}

Since the classification rule only depends on the observed data and the true state transition matrix $A$, a reliable empirical estimation of $A$ can be used in the classifier $\mathcal{C}$ to distinguish normal operations from previously unseen faulty operations. The process of transferring a learned classifier from one building to another is outlined in Alg.~\ref{alg:transfer}. We use data collected from building 1, $X^{(1)} = [x_{1}, x_{2},\ldots, x_{T}]$ and $S^{(1)} = [s_{0}, s_{1},\ldots, s_{T-1}]$ generated by a building which has been certified to be running or simulated at normal operations to estimate $\hat{A}_{1}$ by solving the least squares problem:
\begin{equation}\label{eqn:kernreg}
    \hat{A}_{1} = \underset{W}{\textnormal{argmin}} || WS^{(1)} - X^{(1)} ||_{2}^{2}
\end{equation}

To transfer the classifier to an architecturally similar building 2, a new set of samples $X^{(2)}$ and $S^{(2)}$ is collected from the building. Rather than solving \eqref{eqn:classrule} over again, however, we use the previously learned $\hat{A}_{1}$ as the initial value when solving the new kernel regression problem with weighted least squares \cite{ruppert1994multivariate} (WLS) to find the new estimation of the state transition matrix $\hat{A}_{2}$. Data from building 1, $S^{(1)}$ and $X^{(1)}$ are given low weight, and new data $S^{(2)}$ and $X^{(X)}$ are given higher weight. The building 1 data set serves to constrain the degrees of freedom of the new model, while the building 2 data set updates the operating levels (e.g. temperatures, power consumption) of the various HVAC components under consideration. 

Because the thermodynamic laws that govern the HVAC systems in building 1 and building 2 are the same---only the climate, operating characteristics of the HVAC components, and the building materials may differ---we assume the span of $A_{1}$ and $A_{2}$ to be similar sets; thus initializing gradient descent at $\hat{A}_{1}$ to learn $\hat{A}_{2}$ will require fewer samples from building 2. Cross-validation is used to determine optimal choices of weights for WLS.

\begin{algorithm}%[t!]
\SetAlgoLined
\KwData{Building 1 data $(S^{(1)}, X^{(1)})$; building 2 data $(S^{(2)}, X^{(2)})$}
\KwResult{State estimators $\hat{A}_{2}$, $\mathcal{C}(s, x, \hat{A}_{2})$}
 
  split train/validate data sets for building 2 $(S^{(2)}, X^{(2)})$\;
 
  minimize WLS problem $\hat{A}_{2} = \textnormal{argmin}_{W}|| W[S^{(1)},S^{(2)}] - [X^{(1)},X^{(2)}] ||_{2}^{2}$\;
  
  cross-validate optimal weights with validation data set for building 2
  
  \Return{$\mathcal{C}(s, x, \hat{A}_{2})$}

\caption{Algorithmic outline of learning and transferring state estimation and fault classifier between two buildings}\label{alg:transfer}
\end{algorithm}

This algorithm requires that building 1 and building 2 have comparable HVAC systems, where each system component represented in a row or column in $\hat{A}_{1}$ has an analogous entry (possibly aggregated, e.g. sum of supply and exhaust fan power) in the true state transition matrix $A_{2}$ we wish to transfer our estimate $\hat{A}_{1}$ to. Note that this requirement can be relaxed for if only a subset of components are of interest. That is, building 1 only needs to have similar components to those we wish to detect faulty operation for in building 2.

\section{RESULTS}
\label{sec:results}
Here we state how the classifier $\mathcal{C}$ is computed from the probabilities found in \eqref{eqn:classeval}. Then we present illustrative numerical results on the number of additional samples required to transfer a classifier based on a matrix $\hat{A}_{1}$ learned from data simulated by EnergyPlus to a new matrix $\hat{A}_{2}$ based on data from PNNL's SEB. Then we conclude with transferring a classifier designed to detect EnergyPlus simulated degradation in a variable air volume (VAV) box supply fan due to a fouling filter from an office building operating in a Seattle winter climate to a similar building operating in a Seattle summer climate.\footnote{All code and EnergyPlus configurations used for this paper are available in our Git repository (\url{github.com/cpatdowling/building_transfer/notebooks/build_sys_2019_current.ipynb}).}

\subsection{Posterior probability of fault matrix $\tilde{A}$}

In order to derive the equation found in \eqref{eqn:classeval}, we need to compute \eqref{eqn:posterior}: the posterior probability of the fault matrix $\tilde{A}$. An $n \times p$ matrix normal random variable $Z \sim \mathcal{N}(M, U, V)$ centered at $M$ has a PDF of the form

\begin{subequations}
\begin{align}
    P(Z|M,U,V) &= \frac{1}{(2\pi)^{np/2}|V|^{n/2}|U|^{p/2}}  \nonumber\\ 
    & \qquad \boldsymbol{\cdot} e^{-\frac{1}{2}Tr\left[ V^{-1}(X- M)^{T}U^{-1}(X - M) \right]}\label{eqn:multinormal}
\end{align}
\end{subequations}
This is a random matrix where each element $z_{i,j}$ is normally distributed around $m_{i,j}$, with row-wise covariance matrix $U$ ($n \times n$) and column-wise covariance matrix $V$ ($p \times p$). If we assume the covariance matrices $U$ and $V$ are identity, then each element of $Z$ is independent of the others. 

With an abuse of notation on the indefinite integral, a matrix normal prior on $\tilde{A}$ with identity row- and column-wise covariance, the conditional probability of observed $x$ given $s$ is given by,

\begin{subequations}
    \begin{align}
        P( x | \tilde{A}, s ) &= \int \frac{1}{(2\pi)^{n/2}}e^{-\frac{1}{2}Tr\left[ (x - Bs)^{T}(x - Bs) \right]} \nonumber \\
        &\qquad \boldsymbol{\cdot} \frac{1}{(2\pi)^{np/2}}e^{-\frac{1}{2}Tr\left[ (B - A)^{T}(B - A) \right]} \partial B.
        \label{eqn:posterior} 
    \end{align}
\end{subequations}

We can compute this probability by combining the exponential terms, completing the square, and rescaling the integrand such that the integral evaluates to 1 by definition. We therefore have that \eqref{eqn:posterior} equals

\begin{subequations}
\begin{align}
    & Tr\left[(x - Bs)^{T}(x - Bs) + (B-A)^{T}(B - A)\right] \\
    & = Tr\left[ B^{T}Bxx^{T} + B^{T}B - 2B^{T}yx^{T} - \right. \nonumber \\
    & \qquad \qquad \left. 2B^{T}A + yy^{T} + AA^{T}\right] \\
    & = Tr\left[ B^{T}B(ss^{T} + I) - 2B^{T}(xs^{T} + A) \right. \nonumber \\
    & \qquad \qquad \left. + (xx^{T} + AA^{T}) \right],
\end{align}
\end{subequations}

\noindent and results in a square in terms of $B$. Letting $p \times p$ matrix $C$ and $n \times p$ matrix $D$ defined as \eqref{eqn:C} and \eqref{eqn:D} respectively we have that

\begin{subequations}
\begin{align}
    & Tr\left[ B^{T}BC - 2B^{T}D \right] + Tr\left[(xx^{T} + AA^{T})\right] \nonumber \\
    &\qquad + Tr\left[(xx^{T} + AA^{T})\right] \\
    &= Tr\left[ C(B - DC^{-1})^{T}(B - DC^{-1}) \right] \nonumber \\
    &\qquad - Tr\left[ (DC^{-1})^{T}D \right] + Tr\left[(xx^{T} + AA^{T})\right] \label{eqn:finalsquare}
\end{align}
\end{subequations}

Note that the first trace term in \eqref{eqn:finalsquare} contains the only term with $B$. We can factor the second and third trace terms out of the integrand in \eqref{eqn:posterior}. Indeed,

\begin{subequations}
\begin{align}
    P(x|\tilde{A},s) &= \frac{1}{(2\pi)^{n/2}} e^{-\frac{1}{2}\left(- Tr\left[ (DC^{-1})^{T}D \right] + Tr\left[ xx^{T} + AA^{T} \right]\right)} \nonumber \\
    & \boldsymbol{\cdot} \int \frac{1}{(2\pi)^{np/2}}e^{-\frac{1}{2} Tr\left[ C (B - DC^{-1})^{T}(B - DC^{-1}) \right]} \partial B 
\end{align}
\end{subequations}

\noindent Let,

\begin{equation}
H:= \frac{1}{(2\pi)^{n/2}} e^{-\frac{1}{2}\left(- Tr\left[ (DC^{-1})^{T}D \right] + Tr\left[ xx^{T} + AA^{T} \right]\right)}.
\end{equation}

\noindent $C$ is analagous to the column-wise covariance $V$ size $p \times p$. Noting that $C$ is always invertible and also size $p \times p$,  we re-scale the integral by $|C^{-1}|^{p/2}$ such that it evaluates to 1 (by definition in \eqref{eqn:multinormal}), thus we have that,

\begin{subequations}
\begin{align}
    & H \int \frac{1}{(2\pi)^{np/2}}e^{-\frac{1}{2} Tr\left[ C (B - DC^{-1})^{T}(B - DC^{-1}) \right]} \partial B \\
    &= H|C^{-1}|^{p/2} \int \frac{1}{(2\pi)^{np/2}|C^{-1}|^{p/2}} \nonumber \\
    &\qquad \qquad \qquad \boldsymbol{\cdot}e^{-\frac{1}{2} Tr\left[ C(B - DC^{-1})^{T}(B - DC^{-1}) \right]} \partial B\\
    &= H|C^{-1}|^{p/2} \cdot 1.
\end{align}
\end{subequations}

\noindent Thus we have the probablility that the observed samples were not generated by the true state transition matrix $A$,

\begin{equation}
    P(x|\tilde{A},s) = \frac{1}{(2\pi)^{\frac{n}{2}}} e^{-\frac{1}{2}Tr\left[ ( xx^{T} + AA^{T} - (DC^{-1})^{T}D \right]} |C^{-1}|^{\frac{p}{2}}.\label{eqn:faultprob}
\end{equation}

\subsection{Classifier}

Combining probabilities \eqref{eqn:nofaultprob} and \eqref{eqn:faultprob}, we can use the difference of their respective log-likelihoods to derive the classifier $\mathcal{C}$ by substituting the computed probabilities into \eqref{eqn:classrule}. For classifier values greater than 0, an observation $(s, x)$ is more likely to have come from an HVAC system operating normally, while values less than 0 indicates otherwise. Simplifying gives us the state classification rule in \eqref{eqn:classeval}.

%\begin{subequations}
%\begin{align}
%    0 &\lessgtr Tr\left[ (x - As)^{T}(x - As)\right] - Tr\left[xx^{T} + AA^{T} - %C^{-1}D^{T}D \right] \nonumber \\
%    &\qquad - p\log(|C^{-1}|) 
%\end{align}
%\end{subequations}

The joint probability of output values $x_{1},\ldots, x_{T+1}$ conditioned on the input values $s_{0},\ldots, s_{T}$ is independent since the noise is i.i.d. 
% since each pair depends only on the realization of the IID noise $\epsilon_t$ at each time. 
This means the log-likelihood is additive, and for multiple samples, \eqref{eqn:classrule} can be written as,

\begin{equation}
    0 \lessgtr \sum_{t = 0}^{T} \left[ \log [ P( x_{t+1} | A, s_{t} ) ] - \log [ P( x_{t+1} | \tilde{A}, s_{t} ) ] \right],\label{eqn:sumclass}
\end{equation}

\noindent and a sequence of observations $(S,X)$ can be used to increase the confidence of the classifier $\mathcal{C}$. This assumes that both the noise in the data \emph{and} the degree of perturbation in faulty state transitions $\hat{A}$ has unit variance. This is not the case in our EnergyPlus simulations, let alone in practice. To overcome this, in Sec.~\ref{sec:classtransfer} we use the logdet term and separated trace terms of \eqref{eqn:classeval} as input features of a logistic regression classifier; practically we find and demonstrate below that this allows us to momentarily sidestep the problem of estimating the variance of the elements of $\hat{A}$ when subject to the occurrence of a fault, but demonstrates the validity of the terms of \eqref{eqn:classeval} as being the correct featurization of data $(S,X)$ and estimated $\hat{A}$ for naive fault classification. We will denote this modification of the classifier as $\mathcal{C}_{\log}$.  %Intuitively, \eqref{eqn:sumclass} uses the projection terms to classify whether or not the principal components of the output sequence $X$ is likely to be the basis of $A$. 

\subsection{Numerical Results: Classification}

As an initial demonstration of the effectiveness of $\mathcal{C}$ via Monte Carlo, we consider a fixed $2 \times 2$ matrix $A$ with diagonal entries $0.9$ and $-0.4$ and $0$ elsewhere. For each Monte Carlo iteration, we  perturb the entries of $A$ such that $a_{i,j} \sim \mathcal{N}(a_{i,j},1)$ to generate an unseen faulty operations matrix $B$. We then generate 1000 IID samples of input-output pairs $(S_{A}, X_{A})$ and $(S_{B}, X_{B})$ using \eqref{eqn:statetransition} where input values are also normally distributed with zero mean and unit variance.

\begin{figure}
    \centering
    \includegraphics[width=0.75\columnwidth]{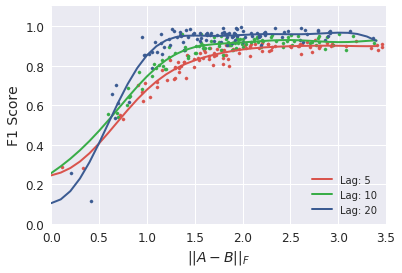}
    \caption{F1 score versus divergence of Monte Carlo realizations of faulty operations state transition matrix $B$, when classifying with an increasing number of sample pairs (``lag'')}
    \label{fig:numericalf1}
\end{figure}

Using \eqref{eqn:sumclass} as our classification rule $\mathcal{C}$, we compute the F1 score for increasing numbers of sample pairs (``lag'') per classification. Figure \ref{fig:numericalf1} illustrates the F1 score as a function of $||A - B||_{F}$ for each realization of $B$, with simple lines of best fit to illustrate the trend. As the Frobenius norm of the difference between $A$ and $B$ increases the the bases of $A$ and $B$ are more likely to diverge and span increasingly different sets. Once more than a very small difference between $A$ and $B$ emerges, the F1 score of $\mathcal{C}$ approaches 1. %Further, increasing the number of points considered improves the performance of the classifier.

%\begin{figure}
%    \centering
%    \includegraphics[width=0.7\columnwidth]{figs/numerical_recall.png}
%    \caption{Caption}
%    \label{fig:my_label}
%\end{figure}

\subsection{Transfer Results: Feasibility and samples savings performance}

Here we demonstrate that a state transition model learned on one building can be transferred to another with similar HVAC system components using a limited number of samples. Building 1 is simulated with EnergyPlus (v. 9.1.0) using a pre-configured example medium office building and Seattle, WA weather data. Building 2 is the SEB on PNNL's main campus in Richland. For both buildings we collect hourly total building power demand, primary air handling unit (AHU) supply and exhaust fan power demand, total lighting power demand, main floor internal zone temperature, as well as outdoor temperature, humidity, day of week and hour of day. All training data is normalized by feature according to the minimum and maximum training data values and time data is embedded as coordinates on a unit circle.

Each building differs in a small number of ways: the medium office building configuration in EnergyPlus is 3 floors, with three AHU's and 3 VAV boxes---1 VAV box per floor; hot and chilled water are provided for on-site. The SEB building has 23 VAV boxes served by two AHU's and 2 additional AHU's with constant-speed fans serving one constant air volume box each. For the purpose of the transfer we select a single, centrally located VAV box on the 1st floor of the EnergyPlus simulated medium office building, and on the SEB main floor (VAV-100), and take the indoor zone temperature measured near each central VAV box. Hot water for SEB is provided for on-site but chilled water is supplied by a central campus plant\footnote{Further details about and floor plan of the building can be found in \cite{dong2019online}.}. Also, significantly, the climate in Seattle is characteristically cool and wet while Richland is located at a higher altitude, arid plateau and sees far greater temperature variation. EnergyPlus data was generated by a full year's simulation and data for SEB was collected from the months of March-October. Both chronologically ordered sets were split into 50/50 train/validation, exhibiting different operational patterns between each split, such as increased power usage during the summer months for cooling. 

\begin{figure}
    \centering
    \includegraphics[width=0.8\columnwidth]{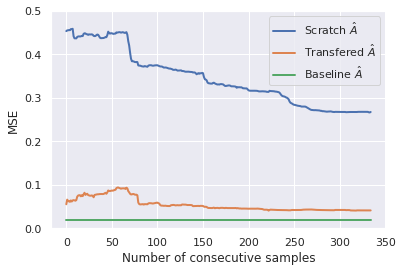}
    \caption{Validation MSE of SEB's $\hat{A}_2$ state transition matrix learned with an increasing number of consecutive hourly samples (normalized in 0-1 over training data) with (``transfer'') and without (``scratch'') including data from EnergyPlus simulations of building 1}
    \label{fig:transferlinear}
\end{figure}

We learn a  state transition matrix for building 1, $\hat{A}_1$, using 6 months (January-June) of operations training data simulated by EnergyPlus. Cross-validation for the selection of the polynomial degree parameter ($d = 4$), as well as a weight regularization parameter ($\alpha = 0.5$) on the Frobenius norm of $\hat{A}_{1}$, was performed with 6 months of validation data (July-December). Using identical model parameters, a state transition matrix $\hat{A}_{2}$ Fig.~\ref{fig:transferlinear} illustrates the averaged mean squared error (MSE) of $\hat{A}_{2}$ when trained on an increasing number of consecutive hourly samples both with the full training data set from building 1 (``Transfered $\hat{A}$'') and without data from building 1 (``Scratch $\hat{A}$''). The baseline in Fig.~\ref{fig:transferlinear} is the best possible validation performance when $\hat{A}_{2}$ is trained on the full SEB training data set.%, as the least squares solution to,

%\begin{equation}\label{eqn:linreg}
%\begin{aligned}
%\left[\begin{array}{c}
%    \textnormal{bldg. power}\\
%    \textnormal{fan power}\\
%    \textnormal{zone temp}\\
%\end{array}\right]_{t+1} = A \cdot \left[\begin{array}{c}
%    \phi_{4}(\textnormal{\tiny{bldg power}})\\
%    \phi_{4}(\textnormal{\tiny{fan power}})\\
%    \phi_{4}(\textnormal{\tiny{in. temp}})\\
%    \phi_{4}(\textnormal{\tiny{out. temp}})\\
%    \phi_{4}(\textnormal{\tiny{out. humidity}})\\
%    \phi_{4}(\textnormal{\tiny{weekday}})\\
%    \phi_{4}(\textnormal{\tiny{hour of day}})\\
%     \phi_{4}(\textnormal{\tiny{lights power}})\\
% \end{array}\right]_{t} + \alpha ||A||_{F}
% \end{aligned} 
% \end{equation}

% and using building 1 model parameters. 

In each case---transfer and scratch---100 training sequences of SEB data two weeks in length were randomly sampled from the training data set (March-June) and the validation data (July-October) performance were averaged across training instances. The dramatically increased performance of the transferred $\hat{A}$ given approximately 3 days worth of samples stems the inclusion of data from building 1 constraining the degrees of freedom of the model by using WLS (with weights $0.01$ and $10.0$ for building 1 and SEB data respectively). The transferred model achieves an MSE of 0.041 (on 0-1 normalized data) vs an MSE of 0.267 for a model learned from scratch, with the best possible performance of an MSE of 0.019 on validation data for a model trained on all SEB training data.

As a comparison, the same procedure is then used to learn a simple, feed-forward neural network $\bd{F}$ with 1 hidden layer twice the dimension of the input equipped with a sigmoid activation function to demonstrate the transfer procedure is not implicit to the usage of weighted least squares or linear methods. $\bd{F}$ is trained with respect to MSE loss on the same input/output pairs, data, and featurization, with a hidden layer twice the dimension of the input. A neural network (NN) $\bd{F}_{1}$ is learned for building 1, and is transferred to a model $\bd{F}_{2}$ of SEB by initialized SGD at the weights learned in $\bd{F}_{1}$ and by using the same data selection procedure as the WLS transfer method. We achieve near baseline MSE with two days worth of samples when transferring, and notably better validation performance overall due to using a more complex model than a single matrix. While more accurate at predicting future states on the same data sets, this neural network lacks intepretability and compatibility with our naive classifier \eqref{eqn:classrule}.

\begin{figure}
    \centering
    \includegraphics[width=0.8\columnwidth]{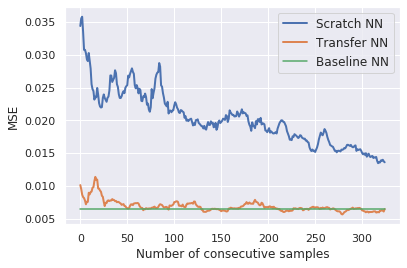}
    \caption{Validation MSE of SEB's $\bd{F}_{2}$ neural network learned with an increasing number of consecutive hourly samples (normalized in 0-1 over training data) with (``transfer'') and without (``scratch'') including data from EnergyPlus simulations of building 1}
    \label{fig:transfernn}
\end{figure}

\subsection{EnergyPlus Simulation Results: Transfer}\label{sec:classtransfer}

We again use EnergyPlus to simulate true HVAC system operations for a pre-configured medium office building. For normal operations during weekday working hours, we learn a state transition matrix $\hat{A}$ that maps input $s_{t}$ containing only outdoor and indoor temperature (on each of 3 floors), outdoor humidity, and VAV box supply fan power consumption (3 fans for 3 VAV boxes total) for 8 consecutive hours and polynomial kernel $\phi_{2}$ to single output $x_{t+1}$ of VAV box supply fan power consumption in the next hour. 

First we simulate the office building for an entire year in typical Seattle weather. $\hat{A}$ is computed by solving the least squares problem \eqref{eqn:kernreg} using normal operations data only. Using EnergyPlus' built-in fault simulation, we then simulate a single VAV box supply fan's filter in the building as becoming increasingly fouled \cite{zhang2016modeling}\footnote{EnergyPlus Version 8.9.0 Documentation
Engineering Reference Section 11.2.4 'Air Filter Fouling' \url{https://energyplus.net/sites/all/modules/custom/nrel_custom/pdfs/pdfs_v8.9.0/EngineeringReference.pdf}} each \emph{full} week (resulting in up to +20\% of normal air flow resistance) over the same year's weather data. The true and VAV box supply fan fault state transition matrices have very similar bases, so a much larger 140 consecutive hourly sample pairs during normal business hours is required to achieve sufficiently accurate classification results when training.

\begin{figure}
    \centering
    \includegraphics[width=0.8\columnwidth]{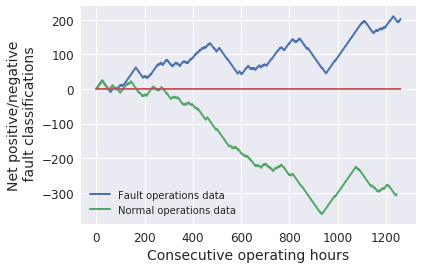}
    \caption{Classification validation performance for $\hat{A}$ for a simulated medium office building learned on a full year of operations in Seattle.}
    \label{fig:netclass}
\end{figure}

The row and column-wise covariance of the true, normal operations state transition matrix $A$ are not identity. Without knowing how the covariance in the simulation data scales the terms of the traces in \eqref{eqn:classeval}, we use these terms as inputs to a logistic regression classifier $\mathcal{C}_{\log}$ which we then train after solving for $\hat{A}$ on normal, non-faulty operations data. For a full year of simulated normal and fault building operations, without covariance information $\mathcal{C}_{\log}$ achieves a 58.8\% and 56.13\% validation precision and recall using \emph{only} indoor, outdoor temperature, humidity, and fan power consumption data on individual samples. Fig.~\ref{fig:netclass} illustrates net fault (+1) vs. no fault (-1) classifications for a continuous sequence of simulated validation data.

To test transferring a state transition matrix $\hat{A}_{1}$ from an office building operating in a Seattle winter climate $A_{1}$ to a building operating in a Seattle summer climate $A_{2}$. We solve \eqref{eqn:kernreg} with only two weeks of normal operation data in the winter to find $\hat{A}_{1}$. When tested on two weeks of winter validation fault/no fault data, $\mathcal{C}_{\log}$ trained on two weeks of winter fault/no fault data achieves a precision of 55.15\% and recall of 54.59\%. To transfer $\hat{A}_{1}$ to $\hat{A}_{2}$ we use two weeks of summer normal operations data, and following Alg.~\ref{alg:transfer}, initialize \eqref{eqn:kernreg} with $\hat{A}_{1}$ and apply WLS to compute $\hat{A}_{2}$. $\mathcal{C}_{\log}$ trained on two weeks of summer fault/no fault data achieves a precision of 56.37\% and recall of 58.67\% on summer validation data according to eqn.~\eqref{eqn:sumclass}.

\section{CONCLUSION}
\label{sec:conclusion}
In this work we have demonstrated a novel, \emph{transferable} Bayesian classifier for detecting faults due to HVAC component degradation. We employed a matrix normal prior distribution on the grounds that if a linear, time-dynamic system described by a matrix $A$ under normal operations begins to fail, the failure process is generated by an unseen matrix $\tilde{A}$. Our classifier depends only on $A$ and the observed data, and transferring via weighted least squares to a new building is sample efficient. Future work in this direction is rich: utilizing informed priors in the classifier with system knowledge, known failure rates, and fault rules; accounting for differences in empirical covariance between buildings to eliminate the need of a logistic layer around $\mathcal{C}$ and thus the requirement for sample fault data; and learning control theory compatible state transition matrices $\hat{A}$ via, for example, regularization and eigenvalue constraints.

%\balance
\bibliographystyle{IEEEtran}
\bibliography{main}

\end{document}